\documentclass[10pt,twocolumn,letterpaper]{article}

\usepackage{cvpr}              

\definecolor{bestcolor}{RGB}{222,235,247}   
\definecolor{secondbestcolor}{RGB}{249,233,222}   
\definecolor{lightblue}{HTML}{4B91C0}
\usepackage{colortbl}
\usepackage{booktabs}
\usepackage{multirow}
\usepackage{capt-of}
\usepackage{flushend, cuted}
\definecolor{lightgray}{HTML}{E0E0E0}
\usepackage{bm}

\definecolor{cvprblue}{rgb}{0.21,0.49,0.74}
\usepackage[pagebackref,breaklinks,colorlinks,allcolors=cvprblue]{hyperref}

\usepackage{tcolorbox}
\usepackage{listings}
\usepackage{xcolor}
\definecolor{bgcolor}{rgb}{0.95, 0.95, 0.92} 
\definecolor{bordercolor}{rgb}{0.75, 0.75, 0.75} 
\definecolor{textcolor}{rgb}{0.1, 0.1, 0.1} 
\usepackage{float}

\def\paperID{9095} 
\def\confName{CVPR}
\def\confYear{2025}

\makeatletter
\newcommand{\printfsymbol}[1]{%
\textsuperscript{\@fnsymbol{#1}}%
}
\makeatother

\title{Beyond Pixels: Text Enhances Generalization in Real-World Image Restoration}

\author{
Haoze Sun\textsuperscript{1} \quad Wenbo Li\textsuperscript{2}\printfsymbol{1} \quad Jiayue Liu\textsuperscript{1} \quad Kaiwen Zhou\textsuperscript{2} \quad Yongqiang Chen\textsuperscript{3}\\
Yong Guo\textsuperscript{2} \quad Yanwei Li\textsuperscript{3} \quad Renjing Pei\textsuperscript{2} \quad Long Peng\textsuperscript{4} \quad Yujiu Yang\textsuperscript{1}\thanks{Corresponding author} \\
${^1}$Tsinghua University \quad ${^2}$Huawei \quad ${^3}$CUHK \quad ${^4}$USTC \\
{\tt\small shz22@mails.tsinghua.edu.cn} \quad
{\tt\small fenglinglwb@gmail.com} \vspace{-10mm}
}

\begin{document}
\maketitle

\begin{strip}
\vspace{-3mm}
\centering
\includegraphics[width=0.9\textwidth]{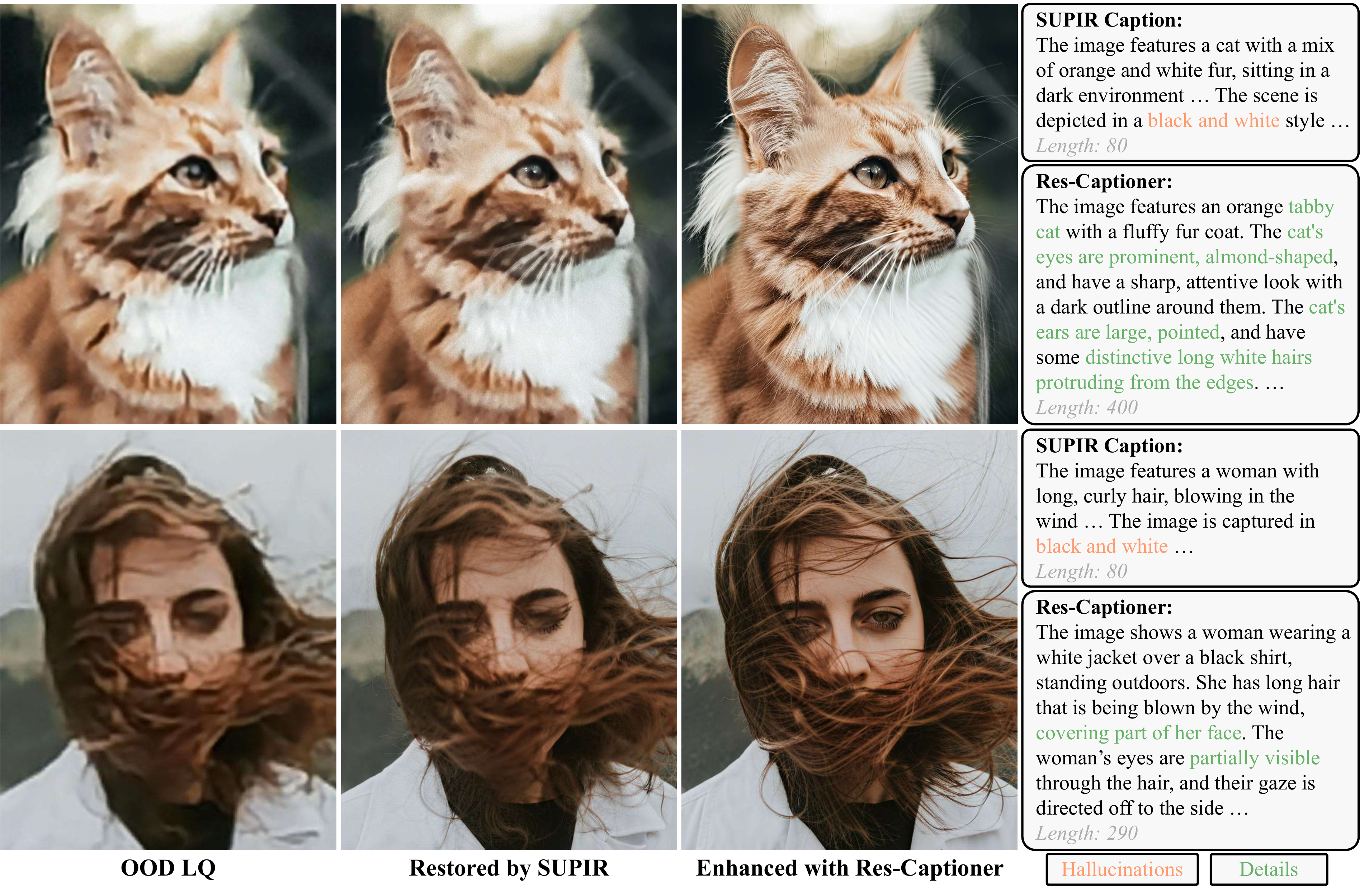}
\captionof{figure}{State-of-the-art methods like SUPIR~\citep{supir} are limited in utilizing their full generative capacity, often yielding blurred or otherwise unsatisfactory results on out-of-distribution (OOD) data, a phenomenon we term as ``generative capability deactivation''. Our Res-captioner can reactivate their generative capabilities by providing detailed and accurate descriptions.}\label{fig:teaser}
\end{strip}

\begin{abstract}
Generalization has long been a central challenge in real-world image restoration. While recent diffusion-based restoration methods, which leverage generative priors from text-to-image models, have made progress in recovering more realistic details, they still encounter ``generative capability deactivation'' when applied to out-of-distribution real-world data. To address this, we propose using text as an auxiliary invariant representation to reactivate the generative capabilities of these models. We begin by identifying two key properties of text input: richness and relevance, and examine their respective influence on model performance. Building on these insights, we introduce Res-Captioner, a module that generates enhanced textual descriptions tailored to image content and degradation levels, effectively mitigating response failures. Additionally, we present RealIR, a new benchmark designed to capture diverse real-world scenarios. Extensive experiments demonstrate that Res-Captioner significantly enhances the generalization abilities of diffusion-based restoration models, while remaining fully plug-and-play.

\end{abstract}

\section{Introduction}

Diffusion-based image restoration methods~\cite{supir, coser, seesr, stablesr, diffbir, pasd, mpp, promptfix, diff-restorer, moe-diffir}, powered by pre-trained text-to-image (T2I) models~\cite{ldm, sdxl}, achieve superior texture and detail recovery compared to GAN-based methods~\cite{bsrgan, real-esrgan, swinir, dasr, femasr, li2022best}. However, these models still face the out-of-distribution (OOD) challenge~\cite{koh2021wilds}, arising from misalignment between training data and real-world test cases. Real-world degradation simulations~\cite{bsrgan,real-esrgan} offer a common mitigation approach, but a domain gap persists~\cite{gen-iqa, wang2024navigating, kong2022reflash}, especially pronounced for device-induced degradations. As depicted in Figure~\ref{fig:teaser}, even state-of-the-art methods struggle to restore fine textures under complex degradations, a limitation we refer to as ``generative capability deactivation''.

We define image restoration as $\bm{x} = \mathcal{R}(\bm{x}_{lq})$, where $\bm{x}$ and $\bm{x}_{lq}$ denote high-quality (HQ) and low-quality (LQ) images, respectively, and $\mathcal{R}$ is the restoration model. To tackle domain generalization, researchers propose learning a cross-domain invariant representation $\bm{z} = \mathcal{G}(\bm{x}_{lq})$~\cite{arjovsky2019invariant, nguyen2021domain, li2022invariant} and then train a prediction network conditioned on $\bm{z}$: $\bm{x} = \mathcal{H}(\bm{z})$. However, learning degradation-invariant representations with strong generalization and minimal information loss remains difficult in image restoration~\cite{blindsr-survey}, as decoupling content from degradation in the image modality is challenging~\cite{chen2024low, tran2021explore, li2022learning}. To address this, we propose transforming LQ images into the text modality using an image captioner $\mathcal{C}$: $\bm{y} = \mathcal{C}(\bm{x}_{lq})$, leveraging recent multi-modal advancements~\cite{llava, llava-v1.5, share-captioner}. This approach offers two advantages: first, in the text modality, degradation-related descriptions $\bm{y}_{deg}$ can be easily separated, leaving the content-related part $\bm{y}_{cont} = \{w \mid w \in \bm{y}, w \notin \bm{y}_{deg}\}$ as a degradation-invariant representation of $\bm{x}_{lq}$. Second, text naturally activates priors in T2I diffusion models, facilitating enhanced texture recovery~\cite{supir, coser, seesr, pasd, promptfix, diff-restorer}.

However, due to significant information compression during the image-to-text transformation, relying solely on $\bm{y}_{cont}$ cannot fully meet the high-fidelity requirements of image restoration tasks. Therefore, we utilize $\bm{y}_{cont}$ as an auxiliary invariant representation in conjunction with the LQ image input, expressed as: $\bm{x} = \mathcal{R}(\bm{x}_{lq}, \bm{y}_{cont})$. In our framework, image restoration is treated as a dual-conditioned image generation problem. Compared to the text input $\bm{y}_{cont}$, the LQ image $\bm{x}_{lq}$ serves as a much stronger condition, being more closely aligned with the final output. However, when the degradation domain of the LQ image shifts, the information that the model can extract from $\bm{x}_{lq}$ largely decreases, leading to the problem of generative capability deactivation (illustrated in Figure~\ref{fig:teaser}). To address OOD data, we propose adaptively enhancing the auxiliary invariant representation $\bm{y}_{cont}$ through our Restoration Captioner (Res-Captioner), compensating for the information loss from $\bm{x}_{lq}$ due to domain shifts.

While several works~\cite{supir, seesr, daclip, daclip-wild, gandikota2024text} have incorporated text to aid in image restoration, they treat it merely as a standard input in line with the original T2I task. These approaches fail to fully investigate how text can enhance generalizability, leading to an underappreciation of its potential role. To close this gap, we begin by identifying two key properties of text input in T2I diffusion-based restoration models: richness and relevance. Richness is primarily reflected in the length of the text; the more detailed the text, the richer the generated textures. Relevance, on the other hand, measures the correlation between the description and the HQ image content, with higher relevance leading to greater fidelity between the restored image and the ground truth. Building on these properties, we establish criteria for optimal auxiliary texts tailored to real-world LQ data and develop Res-Captioner, which is designed to accommodate varying degradation types and image clarity levels. Notably, due to the universal properties we have identified, Res-Captioner can be seamlessly integrated into various restoration models, boosting generalization without the need for retraining.

Finally, given the limitations of current real-world image restoration benchmarks~\cite{realsr, drealsr}, such as the restricted variety of imaging devices, and narrow content diversity, we introduce a new benchmark called \textbf{RealIR}. RealIR encompasses a broader range of degradation sources, clarity levels, and diverse photographic scenarios. Through this benchmark, we demonstrate that our Res-Captioner significantly improves the generalizability of diffusion-based methods, delivering more detailed and high-fidelity restoration results.

The contributions of this paper can be summarized as:
\begin{itemize}
\item We identify the potential of utilizing text as an ancillary invariant representation to enhance generalizability in image restoration, highlighting two key properties—richness and relevance—and their respective impacts on restoration performance.

\item Building on our findings, we develop the Res-Captioner, which generates adaptively enhanced ancillary invariant representations, improving the generalizability of pre-trained diffusion-based restoration models in a fully plug-and-play fashion.

\item We introduce a new restoration benchmark, RealIR, which will be made publicly available, to comprehensively assess generalizability. Using both our benchmark and existing public datasets, we prove the effectiveness of the Res-Captioner across multiple restoration methods.

\end{itemize}

\begin{figure*}[htbp]
    \centering
    \includegraphics[width=\textwidth]{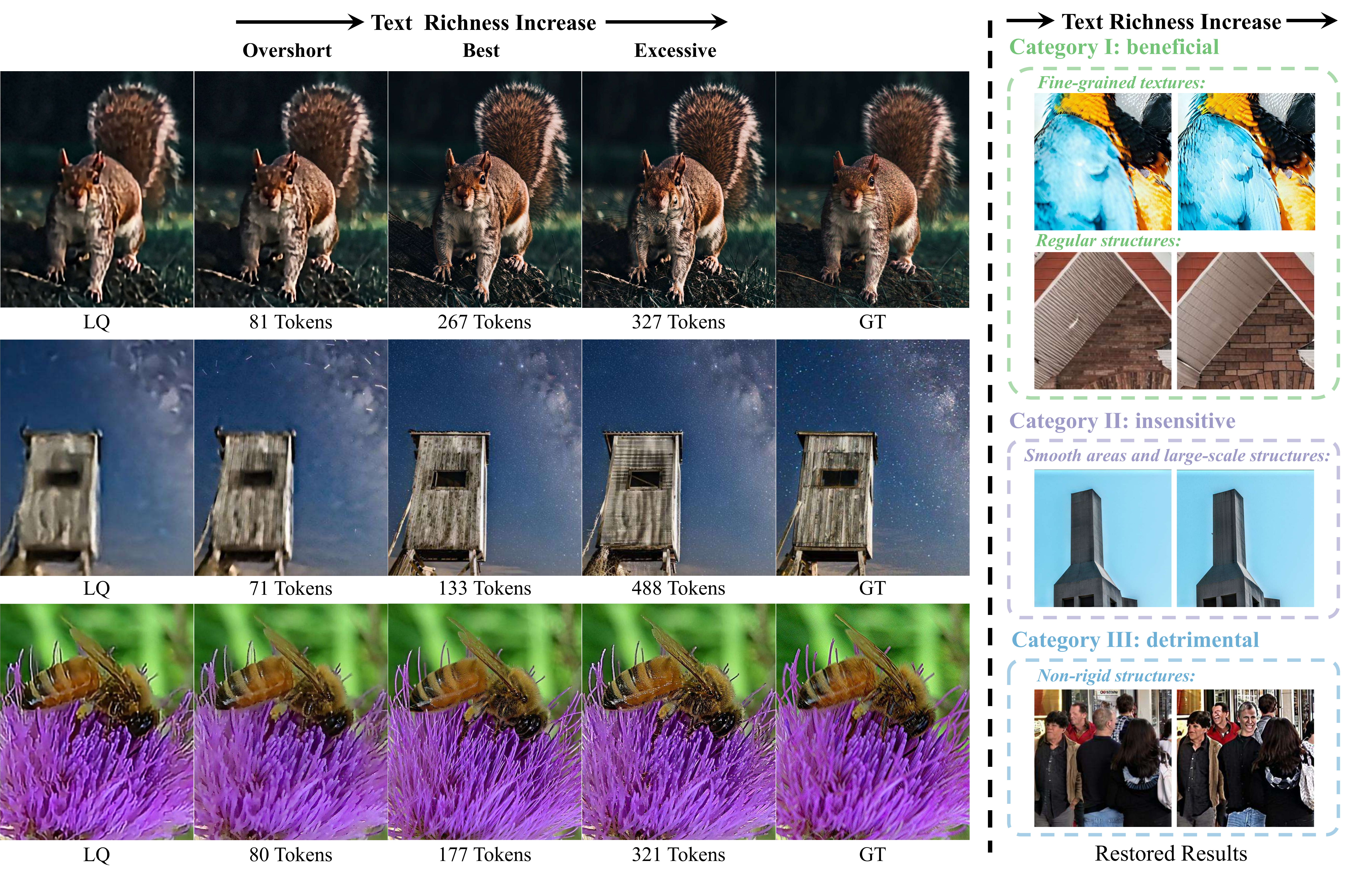}
    \vspace{-7mm}
    \caption{Visualization of the text richness property. (\textbf{Left}) The richness of textures and details in the restored results increases with text richness. Text that is too short can result in the ``generative capability deactivation'' problem.  Excessively long text can lead to messy generation and artifacts. (\textbf{Right}) We can classify image content into three categories based on the effect of increased text richness: \textbf{I} beneficial, \textbf{II} insensitive, and \textbf{III} detrimental.}
    \vspace{-1.5mm}
    \label{fig:richness_image}
\end{figure*}

\section{Restoration Captioner}

\subsection{Properties of Text Input}\label{text_property}


We start by investigating how the text input $\bm{y}$ affects the performance of restoration methods built on text-to-image (T2I) models. We identify two key properties of the text: richness and relevance. Richness refers to the amount of information conveyed, often reflected in text length, while relevance measures the degree of correlation between the text and the corresponding high-quality (HQ) image. Among the properties we observed, text richness emerges as the dominant factor influencing the restored results, which constitutes our core observation.

\subsubsection{Richness Property}
\noindent \textbf{Observation 1}. \textit{The richness of restored textures and details increases proportionally with the text richness.}

\begin{figure}[htbp]
    \centering
    \includegraphics[width=0.48\textwidth]{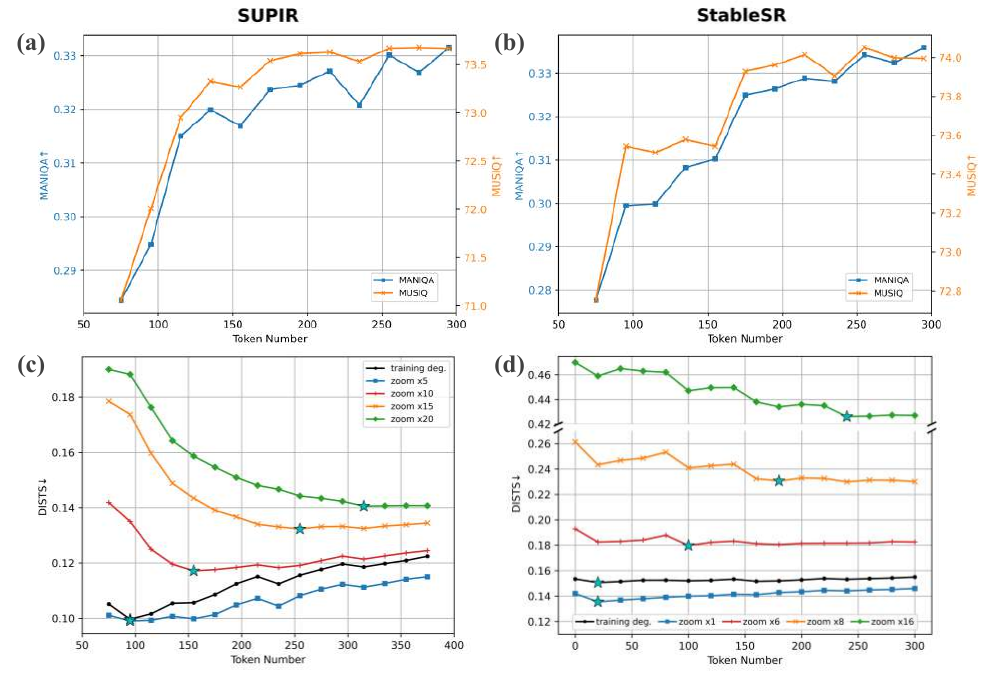}
    \vspace{-6.8mm}
    \caption{Demonstration of the richness property. \textbf{(a, b)}: There is a positive correlation between text richness and the richness of textures in the restored results. \textbf{(c, d)}: The optimal text richness (indicated by an asterisk) is proportional to the degree of deviation between the test degradation domain and the training degradation domain. Best viewed zoomed in.}
    \vspace{-3.8mm}
    \label{fig:richness_line}
\end{figure}

As illustrated in Figure~\ref{fig:richness_image}, we observe that for all low-quality (LQ) images, increasing text richness (\textit{i.e.}, text length) consistently enhances texture restoration.  To explore this further, we prepare a dataset of 120 HQ images from diverse scenarios and generate the corresponding LQ images using Real-ESRGAN~\cite{real-esrgan}. GPT-4 is employed to generate detailed descriptions for the HQ images. We then evaluate two representative restoration models, SUPIR~\cite{supir} and StableSR~\cite{stablesr}. The text input is encoded using CLIP~\cite{clip} to generate 77 tokens. We then repeatedly append the last 20 tokens, excluding the EOS token, and follow~\cite{llmga} to integrate and inject these length-varying tokens into the restoration models to produce the results. Texture richness is assessed using two non-reference metrics: MANIQA~\cite{maniqa} and MUSIQ~\cite{musiq}. As shown in Figure~\ref{fig:richness_line}~\textbf{(a, b)}, both metrics have a positive correlation with the number of text tokens, supporting \textbf{Observation 1}.

We attribute this property to the data bias inherent in pre-trained T2I models, where images with richer content are typically paired with more detailed descriptions during training. Similar observations have been made in T2I research~\cite{dalle3, diffusionmaster}, where longer prompts lead to more enriched scenes. However, in the context of image restoration, this effect primarily enhances texture quality rather than introducing new objects or elements.

\begin{figure*}[t]
    \centering
    \includegraphics[width=\textwidth]{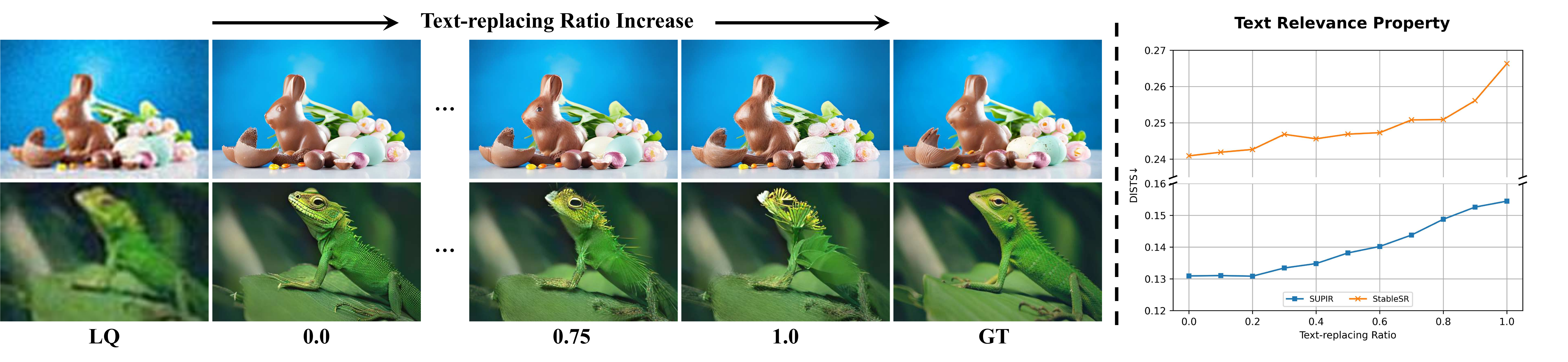}
    \vspace{-7mm}
    \caption{Visualization and demonstration of the text relevance property. \textbf{Left}: The accuracy of textures and details in the restored results decreases as the text-replacing ratio increases, indicating that text relevance contributes to the fidelity of the restoration. \textbf{Right}: DISTS increases with a higher text-replacing ratio, further indicating a decrease in the fidelity of the restored results.}
    \label{fig:relevance}
\end{figure*}

\noindent \textbf{Observation 2}. \textit{The optimal level of text richness is influenced by degradation severity and image content.}


As discussed, detailed text descriptions improve texture restoration. However, as shown in Figure~\ref{fig:richness_image}, exceeding the optimal range of text richness may lead to undesirable artifacts or messy generation. For instance, the squirrel's eyes and mouth are misaligned with the LQ image, and the bee shows over-sharpening effects. We posit that the optimal text richness is proportional to the domain gap between training and testing degradations. To validate this, we prepare LQ images either simulated or captured in the wild with different zoom ratios and evaluate the performance of SUPIR and StableSR in relation to text richness. As illustrated in \ref{fig:richness_line}~\textbf{(c, d)}, as the test degradation increasingly diverges from the training setting (\textit{e.g.}, $4\times$ Real-ESRGAN degradation), the optimal text richness similarly increases. This is because, as degradation severity intensifies, the useful information the model can extract from LQ images diminishes, necessitating more informative textual inputs to compensate for the information loss.

We also find that the optimal text richness depends on the LQ image content. Following~\cite{liang2022details}, we categorize three groups based on the impact of increased text richness: beneficial, insensitive, and detrimental. Category I, ``beneficial'', includes fine-grained textures (\textit{e.g.}, feathers, leaves, sand) and regular structures (\textit{e.g.}, walls, windows), which benefit from longer text input as it activates the model's generative capability. Category II, ``insensitive'', consists of smooth areas and large-scale structures (\textit{e.g.}, sky), where text richness has minimal effect. Category III, ``detrimental'', includes non-rigid structures (\textit{e.g.}, text, crowds), where excessively long text may compromise fidelity.


\subsubsection{Other Properties}\label{sec:harmful_des}
\noindent \textbf{Observation 3}. \textit{The fidelity of restored textures improves in correlation with the relevance of the text description.}

To quantify the text relevance property, we introduce the ``text-replacing ratio'', defined as the ratio of original words in the text input $y$ that are replaced with non-meaningful words like ``the'' or ``for''. As the ratio increases, the relevance between the text and the corresponding HQ image decreases, while the text richness remains unchanged. As shown in Figure~\ref{fig:relevance}, we observe that although restored results retain rich textures with a higher text-replacing ratio, they suffer from a decline in fidelity. This is confirmed by the decreasing DISTS scores (lower is better), measured between the HQ image and the restored output. In contrast to other works~\cite{seesr, supir} that emphasize the text relevance, our observations emphasize the overall relevance of the text input, not just the relevance of the main subjects.

\begin{figure*}[t]
    \centering
    \includegraphics[width=\textwidth]{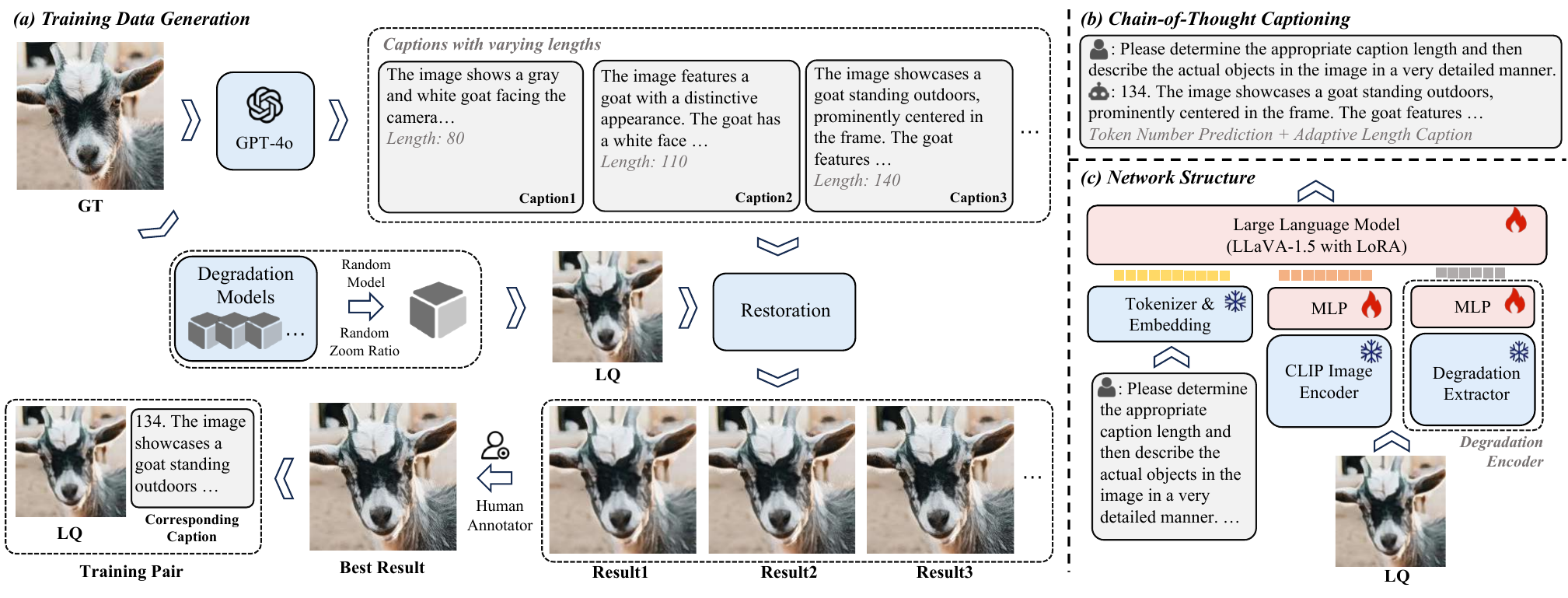}
    \vspace{-7mm}
    \caption{\textbf{(a)} The generation and annotation process of our training data. \textbf{(b)} Chain-of-Thought captioning of our Res-Captioner. \textbf{(c)} Network structure of our Res-Captioner.}
    \vspace{-2mm}
    \label{fig:main}
\end{figure*}

\noindent \textbf{Observation 4}. \textit{Descriptions related to degradation or photography can lead to blurring in the restored images.}

Lin et al.~\cite{lin2024improving} suggest that degradation-related descriptions impair restoration quality. However, we find that photography-related terms, such as ``shallow depth of field'' or ``bokeh effect,'' can also cause blurring in the restored results. Visual examples are provided in the supplementary materials. Additionally, due to the limited spatial control of pre-trained T2I models~\cite{avrahami2023spatext}, even descriptions like ``the background is blurred, while the main subject is sharp,'' which accurately reflect the HQ image, can lead to overall blurring. Thus, it is crucial to eliminate these terms from the text input. While recent works~\cite{supir, coser, seesr, pasd} have introduced negative prompts to improve restoration, they cannot fully counteract the impact of harmful descriptions.




\subsection{Adaptive Caption Learning}\label{sec:training_data}

As discussed in Section~\ref{text_property}, text is crucial in controlling both the richness and fidelity of textures in restored results. However, existing image captioners~\cite{llava, llava-v1.5, share-captioner, hao2024optimizing, hu2024ella}, not designed for restoration, fail to enhance text richness adaptively and may generate harmful descriptions. This contributes to the ``generative capability deactivation'' problem (Figure~\ref{fig:teaser}) in real-world scenarios. To address this, we introduce Res-Captioner, a restoration-specific captioner that generates high-quality text descriptions for real-world LQ images, ensuring adaptive control over both richness and relevance across diverse degradation levels and content.

\noindent \textbf{Training data generation.} We first collect HQ images from~\cite{unsplash}, ImageNet~\cite{imagenet}, and SAM~\cite{sam}. This ensures a selection of rich-content, high-clarity HQ images from diverse scenarios. Next, as shown in Figure~\ref{fig:main}~(a), we leverage five pre-trained latent diffusion models (LDM)~\cite{ldm} to generate LQ images that simulate varying imaging devices and zoom ratios. Training details are in the appendix. We also include a percentage of Real-ESRGAN-generated~\citep{real-esrgan} LQ images. To ensure high relevance while minimizing hallucination, as illustrated in Figure~\ref{fig:main}~(a), we use GPT-4 to generate descriptions of varying lengths for each HQ image. Several prompting techniques, detailed in the appendix, are applied to avoid degradation-related or photography-specific content. These descriptions are fed into the restoration model, producing multiple restored candidates for each LQ image. Human annotators select the optimal text input that provides the best visual result, balancing texture richness and fidelity. In total, we curate 5,500 LQ image-caption pairs for training our Res-Captioner.



It is important to note that although we used a specific restoration model, SUPIR~\cite{supir}, in our training data collection pipeline, Res-Captioner performs effectively across other restoration models in a plug-and-play manner. This demonstrates that the properties we identified as design guidelines for Res-Captioner are highly generalizable. Furthermore, due to differences in generation capacity between T2I backbones, fine-tuning with a small amount of data—approximately 200 pairs—from the restoration model on a new backbone further improves results (shown in the supplementary materials).

\noindent \textbf{Chain-of-Thought captioning.} To generate descriptions with appropriate richness, we enhance the reasoning and decision-making capabilities of Multimodal Large Language Models (MLLM) by adopting the ``Chain of Thought'' (CoT) strategy~\cite{cot} in Res-Captioner. Specifically, as shown in Figure~\ref{fig:main}~(b), the model first predicts the optimal token number before generating the corresponding caption. As demonstrated in Section~\ref{sec:abl_study}, this approach significantly improves the accuracy of the description length.

\begin{table*}[!t]
\resizebox{\linewidth}{!}{
\begin{tabular}{lcccccccccc}
    \toprule
     \multirow{2}{*}{Methods} & \multicolumn{5}{c}{\textbf{RealIR (Cameras)}} & \multicolumn{5}{c}{\textbf{RealIR (Internet)}} \\
    \cmidrule(lr){2-6}\cmidrule(lr){7-11}
     & MUSIQ$\uparrow$ & MANIQA$\uparrow$ & LIQE$\uparrow$ & NIQE$\downarrow$ & CLIP-IQA$\uparrow$ & MUSIQ$\uparrow$ & MANIQA$\uparrow$ & LIQE$\uparrow$ & NIQE$\downarrow$ & CLIP-IQA$\uparrow$ \\
    \midrule
    Real-ESRGAN+~\cite{real-esrgan}      & 58.54          & 0.1784          & 2.425          & 5.049          & 0.4900          & 58.34          & 0.2048          & 2.157          & 5.646          & 0.4458          \\
    DASR~\cite{dasr}             & 53.82          & 0.1487          & 2.208          & 6.038          & 0.4045          & 50.84          & 0.1397          & 1.594          & 6.748          & 0.3290          \\
    CoSeR~\cite{coser}            & 56.91          & 0.1163          & 2.597          & 4.766          & 0.4789          & 66.67          & 0.1842          & 3.822          & 4.042         & 0.5831          \\
    SeeSR~\cite{seesr}            & 70.19          & 0.2138          & 3.768          & 3.705          & 0.6401          & 72.65          & 0.2694          & 4.243          & 3.749          & 0.6706          \\
    StableSR~\cite{stablesr}         & 66.15          & 0.1924          & 3.466          & 4.208          & 0.6345          & 67.66          & 0.2012          & 3.913          & 4.033          & 0.6400          \\
    StableSR w/ Ours & \cellcolor{lightgray}69.28          & \cellcolor{lightgray}0.2389          & \cellcolor{lightgray}3.693          & \cellcolor{lightgray}3.891          & \cellcolor{lightgray}\textbf{0.6956}          & \cellcolor{lightgray}71.64          & \cellcolor{lightgray}0.2690          & \cellcolor{lightgray}4.279          & \cellcolor{lightgray}3.784          & \cellcolor{lightgray}\textbf{0.7031}          \\
    SUPIR~\cite{supir}            & 60.43          & 0.1651          & 2.983          & 4.213          & 0.4793          & 71.94          & 0.2727          & 4.425          & 3.492          & 0.6362          \\
    SUPIR w/ Ours    & \cellcolor{lightgray}\textbf{71.38} & \cellcolor{lightgray}\textbf{0.2543} & \cellcolor{lightgray}\textbf{4.056} & \cellcolor{lightgray}\textbf{3.454} & \cellcolor{lightgray}{0.6235} & \cellcolor{lightgray}\textbf{73.26} & \cellcolor{lightgray}\textbf{0.3055} & \cellcolor{lightgray}\textbf{4.578} & \cellcolor{lightgray}\textbf{3.389} & \cellcolor{lightgray}{0.6749} \\
    \bottomrule
\end{tabular}
}
\centering
\vspace{-2mm}
\caption{Quantitative comparisons on our RealIR benchmark. We highlight \textbf{best} values and \colorbox{lightgray}{results of Res-Captioner-enhanced models}.}\label{tab:realir}
\end{table*}

\begin{table*}[!t]
\resizebox{\linewidth}{!}{
\begin{tabular}{lcccccccccccc}
    \toprule
     \multirow{2}{*}{Methods} & \multicolumn{4}{c}{\textbf{Light Degradation}} & \multicolumn{4}{c}{\textbf{Moderate Degradation}} & \multicolumn{4}{c}{\textbf{Heavy Degradation}} \\
    \cmidrule(lr){2-5}\cmidrule(lr){6-9}\cmidrule(lr){10-13}
     & DISTS$\downarrow$ & LPIPS$\downarrow$ & MANIQA$\uparrow$ & LIQE$\uparrow$ & DISTS$\downarrow$ & LPIPS$\downarrow$ & MANIQA$\uparrow$ & LIQE$\uparrow$ & DISTS$\downarrow$ & LPIPS$\downarrow$ & MANIQA$\uparrow$ & LIQE$\uparrow$ \\
    \midrule
    StableSR         & 0.1791          & 0.3311          & 0.2256          & 3.699          & 0.1864          & 0.3209           & 0.2297          & 3.603          & 0.2181          & 0.4008          & 0.1676          & 3.047          \\
    \multirow{2}{*}{StableSR w/ Ours} 
                      & \cellcolor{lightgray}\textbf{0.1748} & \cellcolor{lightgray}\textbf{0.3271} & \cellcolor{lightgray}\textbf{0.2712} & \cellcolor{lightgray}\textbf{3.733} & \cellcolor{lightgray}\textbf{0.1774} & \cellcolor{lightgray}\textbf{0.3121} & \cellcolor{lightgray}\textbf{0.2614} & \cellcolor{lightgray}\textbf{3.872} & \cellcolor{lightgray}\textbf{0.1993} & \cellcolor{lightgray}\textbf{0.3883} & \cellcolor{lightgray}\textbf{0.2298} & \cellcolor{lightgray}\textbf{3.502} \\
                           & \textcolor{lightblue}{2.4\%} & \textcolor{lightblue}{1.2\%} & \textcolor{lightblue}{20.2\%} & \textcolor{lightblue}{0.9\%} & \textcolor{lightblue}{4.8\%} & \textcolor{lightblue}{2.7\%} & \textcolor{lightblue}{13.8\%} & \textcolor{lightblue}{7.5\%} & \textcolor{lightblue}{8.6\%} & \textcolor{lightblue}{3.1\%} & \textcolor{lightblue}{37.1\%} & \textcolor{lightblue}{14.9\%} \\ 
    \midrule
    SUPIR            & 0.1821          & 0.3444          & 0.2042          & 3.148          & 0.1883          & 0.3473          & 0.2182          & 3.349          & 0.2159          & 0.4106          & 0.1749          & 2.840          \\
    \multirow{2}{*}{SUPIR w/ Ours}    
                      & \cellcolor{lightgray}\textbf{0.1680} & \cellcolor{lightgray}\textbf{0.3178} & \cellcolor{lightgray}\textbf{0.3065} & \cellcolor{lightgray}\textbf{4.011} & \cellcolor{lightgray}\textbf{0.1621} & \cellcolor{lightgray}\textbf{0.3052} & \cellcolor{lightgray}\textbf{0.3294} & \cellcolor{lightgray}\textbf{4.226} & \cellcolor{lightgray}\textbf{0.1873} & \cellcolor{lightgray}\textbf{0.3754} & \cellcolor{lightgray}\textbf{0.3033} & \cellcolor{lightgray}\textbf{3.991} \\
                           & \textcolor{lightblue}{7.7\%} & \textcolor{lightblue}{7.7\%} & \textcolor{lightblue}{50.0\%} & \textcolor{lightblue}{27.4\%} & \textcolor{lightblue}{13.9\%} & \textcolor{lightblue}{12.1\%} & \textcolor{lightblue}{51.0\%} & \textcolor{lightblue}{26.2\%} & \textcolor{lightblue}{13.3\%} & \textcolor{lightblue}{8.6\%} & \textcolor{lightblue}{73.4\%} & \textcolor{lightblue}{40.5\%} \\
    \bottomrule
\end{tabular}
}
\vspace{-2.8mm}
\centering
\caption{Quantitative comparisons between the official model and the Res-Captioner-enhanced model under different degradation levels. We show the \textcolor{lightblue}{improvement percentage} on each metric.}\label{tab:multi_deg}
\vspace{-1mm}
\end{table*}

\begin{table}[htbp]
\resizebox{\linewidth}{!}{
\begin{tabular}{lcccccc}
        \toprule
         \multirow{2}{*}{Methods} & \multicolumn{3}{c}{\textbf{RealSR}} & \multicolumn{3}{c}{\textbf{DRealSR}} \\
        \cmidrule(lr){2-4}\cmidrule(lr){5-7}
         & DISTS$\downarrow$ & LPIPS$\downarrow$ & LIQE$\uparrow$ & DISTS$\downarrow$ & LPIPS$\downarrow$ & LIQE$\uparrow$ \\
        \midrule
        SUPIR         & 0.2660          & 0.3889          & 3.477          & 0.2906          & 0.4741          & 3.655          \\
        SUPIR w/ Ours & \textbf{0.2474} & \textbf{0.3667} & \textbf{4.081} & \textbf{0.2699} & \textbf{0.4409} & \textbf{4.208} \\
        \bottomrule
    \end{tabular}
}
\vspace{-2.8mm}
\centering
\caption{Quantitative comparisons on RealSR and DRealSR datasets. We highlight \textbf{best} values for each metric.}\label{tab:realsr_drealsr}
\vspace{-3mm}
\end{table}

\noindent \textbf{Network structure.} We fine-tune LLaVA-1.5~\cite{llava-v1.5} using low-rank adaptation (LoRA)~\cite{lora} to serve as our Res-Captioner. Since LLaVA is not designed for LQ images, we enhance its ability to detect image degradations by adding a degradation-aware visual encoder, as shown in Figure~\ref{fig:main}~(c). This encoder includes a pre-trained degradation extractor, sensitive to various degradations~\cite{chen2024low, liu2023degae}, and a lightweight adapter for better degradation extraction. The adapter, built from MLP layers, compresses the token count to 1 and then expands it to N tokens (we set $N=36$), enabling focus on global degradation while ignoring spatial variations.

\subsection{RealIR Benchmark}

Current real-world restoration benchmarks~\cite{realsr, drealsr} are limited by a narrow range of degradation types, imaging devices, and content. To overcome these limitations, we introduce RealIR, a new benchmark featuring 152 real LQ images from eight imaging devices (two DSLRs and six phones) with varying zoom ratios. We also include 53 LQ images from the internet to capture network transmission degradations, which differ from device-specific degradations. The dataset spans diverse content, including portraits, animals, plants, and architectural scenes, enabling comprehensive evaluations of restoration methods' generalizability.


\section{Experiments}

\subsection{Implementation Details}\label{sec:imple_details}



{Our Res-Captioner is built on LLaVA-1.5\footnote{https://huggingface.co/liuhaotian/llava-v1.5-13b}. 
We follow the standard LLaVA LoRA settings and train the model with a batch size of 128 for 500 steps on an A800 GPU, using the Adam optimizer~\cite{adam} with a learning rate of $2 \times 10^{-4}$. 


\begin{figure*}[t] 
    \centering
    \includegraphics[width=\textwidth]{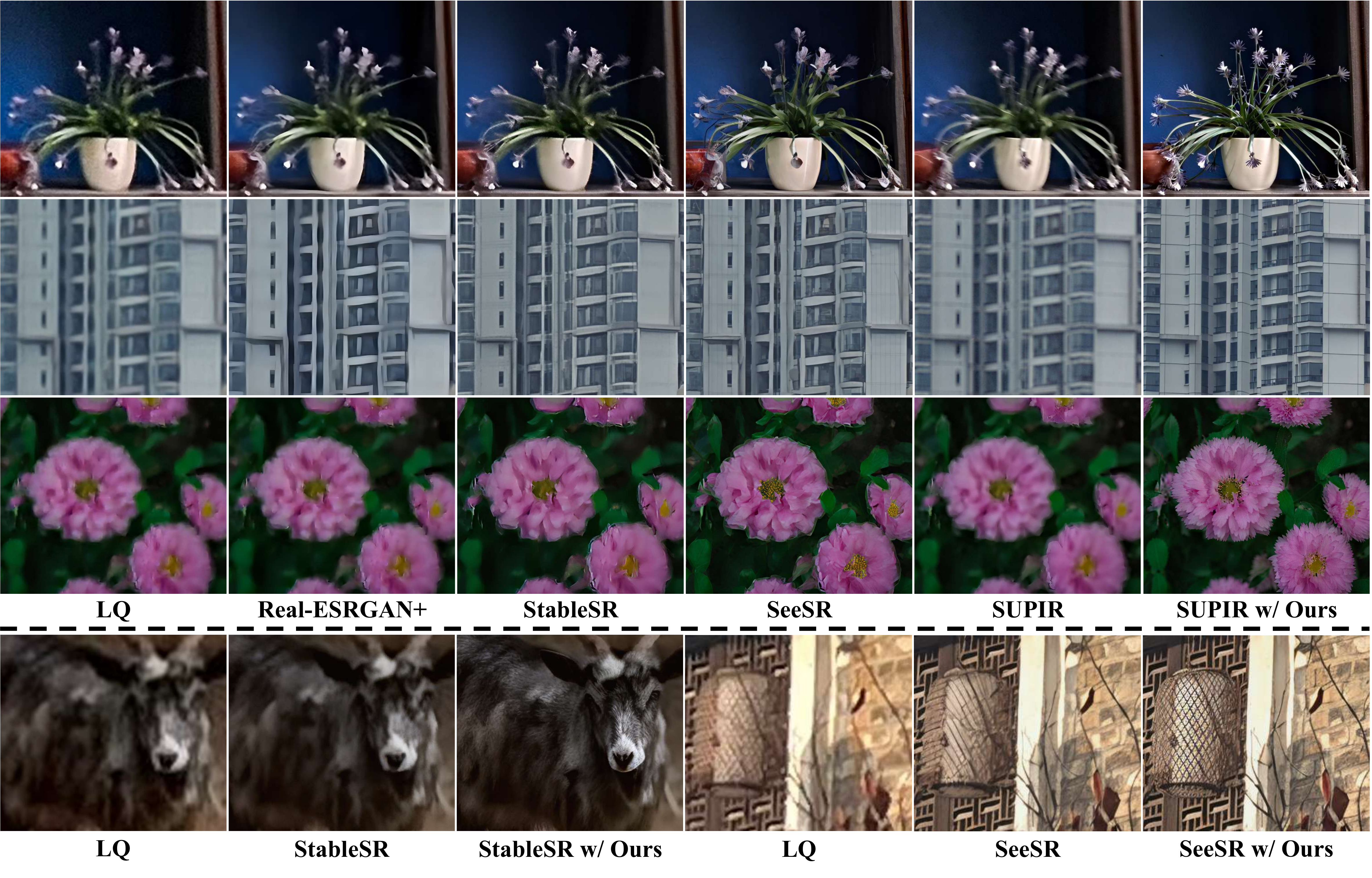}
    \vspace{-6.8mm}
    \caption{Qualitative comparisons on in-the-wild images. \textbf{Upper}: Comparisons between SOTA restoration methods and Res-Captioner-enhanced SUPIR. \textbf{Lower}: Visual quality improvements introduced by Res-Captioner on StableSR and SeeSR.}
    \vspace{-2mm}
    \label{fig:comp}
\end{figure*}

\subsection{Experimental Settings}

\noindent {\textbf{Test datasets.} Our RealIR dataset covers diverse content and degradations from real-world scenes, making it ideal for evaluating restoration models' generalization. Due to the lack of ground-truth images in RealIR, we create an additional multi-degradation test set of 120 LQ-HQ pairs using pre-trained latent diffusion models (LDM). To ensure fair evaluation, the LQ degradations differ from those in the training set. We categorize the pairs into three degradation levels based on zoom ratio: light (3–7), moderate (8–10), and heavy (15–20). These sets enable a thorough evaluation of restoration quality across different degradation levels. Additionally, we assess our approach on established benchmarks like RealSR~\cite{realsr} and DRealSR~\cite{drealsr}, using randomly cropped patches for a more comprehensive analysis.


\noindent {\textbf{Compared methods.} Our experiments include state-of-the-art (SOTA) real-world image restoration methods, such as GAN-based approaches like Real-ESRGAN+~\cite{real-esrgan} and DASR~\cite{dasr}, as well as diffusion-based models including StableSR~\cite{stablesr}, SeeSR~\cite{seesr}, CoSeR~\cite{coser}, and SUPIR~\cite{supir}. 
We integrate Res-Captioners into two representative diffusion-based restoration models: SUPIR~\cite{supir}, and StableSR~\cite{stablesr}. The effect of combining with more models is shown in the supplementary material.


\noindent {\textbf{Evaluation metrics.} For test sets without ground truth, such as RealIR, we use non-reference evaluation metrics aligned with human perception, including MUSIQ~\cite{musiq}, MANIQA~\cite{maniqa}, LIQE~\cite{liqe}, NIQE~\cite{niqe}, and CLIP-IQA~\cite{clip-iqa}. For datasets with ground truth, we adopt perceptual distance metrics like DISTS~\cite{dists} and LPIPS~\cite{lpips}, alongside the LIQE metric, which leverages large vision-language models for robust evaluation. Pixel-level metrics such as PSNR and SSIM are no longer considered, as they exhibit weak correlation with human perception, as discussed in related works~\citep{supir, coser}.}

\subsection{Comparison with State of the Arts}

\subsubsection{Quantitative Results}


{Our quantitative results are organized into two parts. First, we assess the generalization ability of existing restoration methods in real-world scenarios using the RealIR benchmark, showing that Res-Captioner consistently improves their performance. Second, we evaluate the multi-degradation test set and the existing benchmarks, confirming that Res-Captioner improves detail generation while maintaining fidelity across various degradation levels. We also compare other captioners and user-defined prompts to our Res-Captioner in the supplementary materials, highlighting its superior performance.}



\noindent {\textbf{RealIR benchmark.} The results in Table~\ref{tab:realir} evaluate real LQ images from various cameras and the internet. 
Overall, diffusion-based models exhibit superior visual quality compared to GAN-based models, due to their stronger generative capabilities. 
Notably, when integrated with our Res-Captioner, the representative diffusion-based models StableSR and SUPIR show significant improvements across all metrics, proving how our approach fully activates the generative power of T2I-based restoration models for diverse real-world LQ images.}

{The improvement introduced by Res-Captioner varies across different restoration models. For example, Res-Captioner enhances StableSR's LIQE score on manually captured RealIR data by approximately \textbf{6.5\%}, while it increases the SUPIR's score by an impressive \textbf{36\%}. This discrepancy is due to the differing generative capabilities of T2I backbones. SUPIR, which suffered from ``generative capability deactivation'', is effectively reactivated by our Res-Captioner, unlocking the full potential of SDXL~\cite{sdxl}.}

\begin{figure*}[!t]
\centering
\begin{minipage}{0.7\linewidth}
    \resizebox{\linewidth}{!}{
      \begin{tabular}{lcccccc}
        \toprule
         \multirow{2}{*}{Method} & \multicolumn{2}{c}{\textbf{Light Degradation}} & \multicolumn{2}{c}{\textbf{Moderate Degradation}} & \multicolumn{2}{c}{\textbf{Heavy Degradation}} \\
        \cmidrule(lr){2-3}\cmidrule(lr){4-5}\cmidrule(lr){6-7}
         & DISTS$\downarrow$ & LPIPS$\downarrow$ & DISTS$\downarrow$ & LPIPS$\downarrow$ & DISTS$\downarrow$ & LPIPS$\downarrow$ \\
        \midrule
        Ours    & \textbf{0.1680}   & \textbf{0.3178}   & \textbf{0.1621}     & \textbf{0.3052}    & 0.1873            & \textbf{0.3754}   \\
        w/ Min Len.      & 0.1718            & 0.3274            & 0.1753              & 0.3252             & 0.2033            & 0.4009            \\
        w/ Max Len.      & 0.1864            & 0.3525            & 0.1770              & 0.3184             & 0.1964            & 0.4039            \\
        w/ Low Rel. & 0.1738            & 0.3389            & 0.1655              & 0.3061             & 0.1907            & 0.3914            \\
        w/ Harmful Des.  & 0.1686            & 0.3191            & 0.1678              & 0.3178             & \textbf{0.1868}   & 0.3883            \\
        \bottomrule
      \end{tabular}
    }
    \vspace{-3mm}
    \centering
    \captionof{table}{Ablation studies on text richness, relevance, and harmful descriptions. We highlight \textbf{best} values for each metric.}\label{tab:abl_study}
    \vspace{-3mm}
\end{minipage}%
\hfill%
\begin{minipage}{0.285\linewidth}
    \centering
    \includegraphics[width=\linewidth]{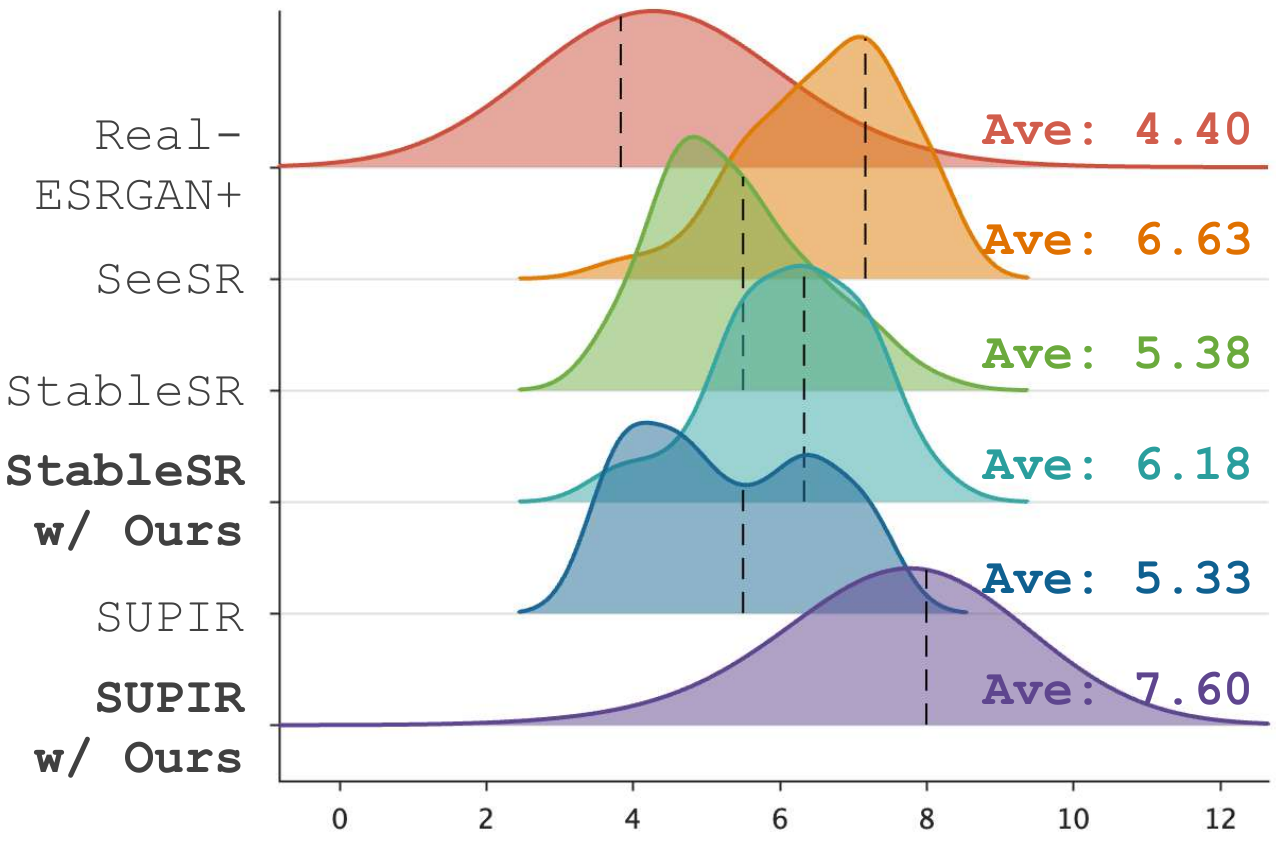} 
    \vspace{-6mm}
    \caption{User study.}\label{fig:user_study}
    \vspace{-4mm}
\end{minipage}
\end{figure*}

\noindent \textbf{Fidelity evaluation.} We compare the original models with their Res-Captioner-enhanced versions on the multi-degradation test set, as shown in Table~\ref{tab:multi_deg}. Res-Captioner consistently improves fidelity for both StableSR and SUPIR, with notable gains in DISTS and LPIPS. Performance improvements increase with the severity of degradation. For instance, the DISTS score of enhanced StableSR improves by approximately \textbf{2.4\%}, \textbf{4.8\%}, and \textbf{8.6\%} for light, moderate, and heavy degradation, respectively. This trend supports our use of text as an auxiliary invariant representation. As test degradation deviates from the training distribution, the restoration model extracts less useful information from the LQ image, making the supplementary text provided by Res-Captioner increasingly beneficial.

{Given the relatively simple and light degradation in RealSR and DRealSR, we use SUPIR as the reference model for evaluation. Our Res-Captioner significantly improves the performance of SUPIR in Table~\ref{tab:realsr_drealsr}, further demonstrating the robustness of our approach in real-world scenarios.}

\subsubsection{Qualitative Results} 



{We provide visual comparisons on in-the-wild LQ images in Figure~\ref{fig:comp}. In the upper section, Real-ESRGAN+ struggles with limited generative capability, failing to restore high-definition textures. Both SUPIR and StableSR experience ``generative capability deactivation'' on out-of-distribution (OOD) data, causing significant blurring. Although SeeSR responds better to OOD data, its textures appear overly smooth and unrealistic. In contrast, Res-Captioner fully activates the generative potential of the T2I backbone in SUPIR, restoring clearer, more realistic textures, such as detailed flower petals and building facades.}


The lower section of Figure~\ref{fig:comp} shows how Res-Captioner enhances other models. Integrated with Res-Captioner, StableSR and SeeSR recover fine-grained textures and structures, like goat fur and lantern mesh, significantly outperforming their original versions. Additional restoration tasks are provided in the supplementary materials.

\subsubsection{User Study}


{To validate Res-Captioner's impact on generalization in real-world scenarios, we conduct a user study with 31 experienced researchers on in-the-wild LQ images. Participants rate the visual quality (1 to 10, where higher is better) of results from Real-ESRGAN+, SeeSR, StableSR, StableSR with Res-Captioner, SUPIR, and SUPIR with Res-Captioner. As illustrated in Figure~\ref{fig:user_study}, both StableSR and SUPIR significantly improve with Res-Captioner, with SUPIR, when enhanced with Res-Captioner, delivering the highest visual quality among all methods.}

\subsection{Ablation Study}\label{sec:abl_study}

{We examine the impact of text properties—richness, relevance, and harmful descriptions—on restoration performance by ablating each in experiments. All models are trained under the same settings, differing only in the training data. We also analyze the effect of our Chain-of-Thought (CoT) captioning and degradation-aware visual encoder on text richness prediction. SUPIR is used as the restoration model in this section.}


\noindent {\textbf{Text richness.} To explore the impact of text richness, we create two training sets with the shortest and longest captions from GPT-4, corresponding to ``w/ Min Len.'' and ``w/ Max Len.'' in Table~\ref{tab:abl_study}. The results show that Res-Captioner (``Ours'') performs best across varying degradation levels, thanks to its adaptive text richness. Additionally, the ``w/ Max Len.'' model outperforms ``w/ Min Len.'' as degradation severity increases, consistent with our \textbf{Observation 2}.}


\noindent {\textbf{Text relevance.} To study this property, we first calculate the length of human-selected optimal captions from GPT-4. We then generate low-relevance captions of the same length using LLaVA-1.5 for training, labeled as ``w/ Low Rel.'' In contrast, Res-Captioner achieves better restoration results, emphasizing the importance of high-relevance descriptions.}


\noindent {\textbf{Harmful descriptions.} In Section~\ref{sec:harmful_des}, we identify harmful descriptions that result in blurring in restored images. Using optimal text richness, we generate captions with these harmful descriptions via GPT-4 and fine-tune Res-Captioner with this data, labeled "w/ Harmful Des." in Table~\ref{tab:abl_study}. The results show that harmful descriptions negatively affect restoration performance, causing an average \textbf{2.7\%} decrease in LPIPS.}


\noindent {\textbf{CoT captioning and degradation-aware visual encoder.} We manually annotate the optimal text length, \(L_o\), for 100 LQ images from the RealIR and multi-degradation datasets. To quantify the prediction error, we define the offset level \(E\) as: \( E = \max\left( \lvert L_o - L \rvert - 15, 0 \right) / 30 \), where \(L\) is the captioner's output length. The mean offset level for our Res-Captioner is 1.27. Without the CoT captioning, the mean offset level increases by \textbf{66.7\%}, and without the degradation-aware visual encoder, it rises by \textbf{31.5\%}. These results highlight the effectiveness of our model's design.}

\section{Conclusion}



We use text as an auxiliary invariant representation to boost the generalizability of T2I-diffusion-based restoration models. By focusing on text richness and relevance, we propose Res-Captioner, which enhances real-world restoration performance in a plug-and-play manner.

\clearpage

\begin{center}
    \Large
    \bf{Supplementary Material}
\end{center}

\renewcommand\thesection{\Alph{section}}
\renewcommand\thesubsection{\thesection.\arabic{subsection}}
\renewcommand\thefigure{\Alph{section}.\arabic{figure}}
\renewcommand\thetable{\Alph{section}.\arabic{table}}

Section~\ref{sec1} summarizes related works. In Section~\ref{sec2}, we present visual examples of our observations and the RealIR datasets, along with details on the generation of our training data. Section~\ref{sec3} includes additional experimental results, such as the improvements achieved through fine-tuning for the new text-to-image (T2I) backbone, comparisons with other captioners and user-defined captions, performance on more restoration models, and further qualitative comparisons on various restoration tasks. Finally, we talk about our limitations and future work in Section~\ref{sec4}.

\setcounter{section}{0}

\section{Related work}\label{sec1}
\subsection{Diffusion-based Image Restoration}
Some prior image restoration models~\cite{ddrm, ddnm, difface, gdp, ssd} use pre-trained diffusion~\cite{guideddiffusion} on ImageNet~\cite{imagenet} as their backbone. However, limited by ImageNet’s constrained data volume, these models struggle to perform well on diverse real-world scenes. Recent works~\cite{supir, coser, seesr, stablesr, diffbir, pasd, mpp, promptfix} often adopt T2I diffusion models (e.g., Stable Diffusion~\cite{ldm}) as their backbone, which produce more realistic textures due to their strong generation capabilities. Despite this, these models still face ``generative capability deactivation'' issues when applied to out-of-distribution data (see Figure~\textcolor{cvprblue}{1} in the main paper). Additionally, semantic information has proven crucial in T2I-based restoration models~\cite{supir, coser, seesr}, but its implementation is not unified. For instance, SUPIR~\cite{supir} uses a pre-trained image captioner, CoSeR~\cite{coser} employs a cognitive encoder to extract semantic data, and SeeSR~\cite{seesr} uses a fine-tuned tag model for semantic input. Our paper is dedicated to two core questions in this field: ``How does semantic information benefit restoration?'' and ``What is the best form of semantic information for image restoration?''

\subsection{Generalization Ability of Image Restoration}
The generalization ability of image restoration models, especially on real-world data, has been a central focus of research. Liu et al.~\cite{liu2023evaluating} introduce a new approach to assess this capability. Some studies~\cite{self1, self2, unsupervised} enhance generalization through self-supervised or unsupervised learning, but despite gains, these methods achieve lower overall restoration quality than supervised training. Another approach involves learning a cross-domain invariant representation. However, achieving strong generalization with minimal information loss remains challenging, as decoupling content from degradation is difficult~\cite{chen2024low, tran2021explore, li2022learning}. As a result, current methods~\cite{du2020learning, li2023learning} struggle to apply effectively to real-world scenes.

\subsection{Image Captioner}
Prior works, such as CoSeR~\cite{coser}, use BLIP2~\cite{blip2} for image captioning; however, its captions are brief and lack sufficient detail for effective restoration. LLaVA~\cite{llava} and LLaVA-1.5~\cite{llava-v1.5} introduced a new era with detailed descriptions, gaining adoption in several restoration models~\cite{supir, promptfix}. More recent open-source visual language models, like ShareCaptioner~\cite{share-captioner} and CogVLM~\cite{cogvlm}, further enhance caption detail and accuracy. However, none of these caption models is specifically designed for restoration. These caption models are trained on high-quality images that are not available for restoration tasks and do not consider properties unique to the restoration task (as shown in Section~\textcolor{cvprblue}{2.1} in the main paper).

\begin{figure*}[htbp]
    \centering
    \includegraphics[width=\textwidth]{harmful_des_v5_compressed.pdf}
    \vspace{-7mm}
    \caption{Harmful descriptions to the image restoration.}
    \label{fig:degdis}
\end{figure*}

\section{Detailed Illustration of our Method}\label{sec2}
\subsection{Observation 4}
\textit{Descriptions related to degradation or photography can lead to blurring in the restored images.}

To validate our observation, we use GPT-4 to generate two captions of similar length: one without harmful descriptions and another including them. To exclude the effects of text richness and relevance, we duplicate the harmless description, labeled ``Without Harmful Description'', and combine both harmless and harmful descriptions to create ``With Harmful Description''. As shown in Figure~\ref{fig:degdis}, the description without harmful terms successfully restores clearer and richer details, while the harmful description leads to global or localized blurred outputs.

\subsection{Details of Training Data Generation}
By reproducing high-definition images, we collect numerous real-world LQ-HQ pairs for training the real-world LQ generation model. Data is gathered from five different devices, and five LQ generation models are trained to represent different types of degradation. We select the latent diffusion model (LDM)~\citep{ldm} as our LQ generator, training it to produce LQ images conditioned on corresponding HQ images. Additionally, the zoom ratio used during image reproduction is incorporated as another part of the conditional information. For each degradation model, we retain one zoom ratio for the generation of multi-degradation test set, and the rest are used to generate the training data.

We use the following prompt to generate captions of varying lengths with GPT-4, while avoiding harmful descriptions through the use of restrictive phrasing.

\begin{figure*}[htbp]
    \centering
    \includegraphics[width=\textwidth]{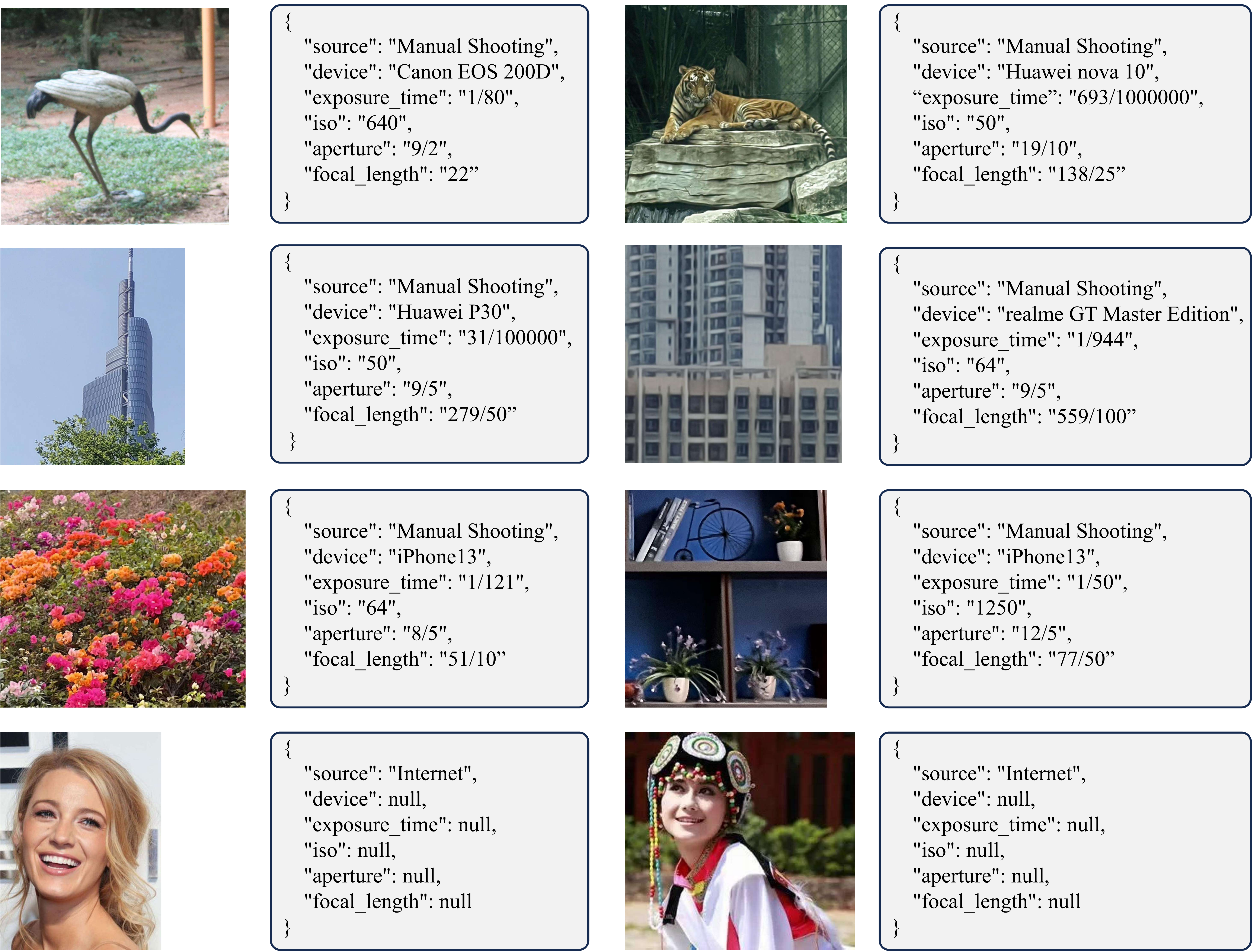}
    \vspace{-7mm}
    \caption{Visualization of RealIR Benchmark.}
    \label{fig:realir_vis}
\end{figure*}

\begin{tcolorbox}[colback=bgcolor, colframe=bordercolor, boxrule=0.5pt, arc=4pt, width=\linewidth, before=\par\vspace{6pt}, after=\par\vspace{6pt}]
\ttfamily
Please describe the actual objects in the image in a very detailed manner. Please do not include descriptions related to the focus and bokeh of this image. Please do not include descriptions like the background is blurred. Please be careful to limit your answer to about XXX words.
\end{tcolorbox}

\noindent We generate a total of seven different caption lengths: 80 words, 110 words, 140 words, 200 words, 260 words, 350 words, and 440 words. The interval between lengths increases progressively, as longer captions tend to cause smaller texture changes when recovering with the same richness interval. The token length of the selected description is then calculated and combined with the description to form the final caption output in the format $\textless$token length, description$\textgreater$ to accommodate our Chain-of-Thought captioning (see Section~\textcolor{cvprblue}{2.2} in the main paper). 

When training and testing the Res-Captioner, we use the following prompt:


\begin{tcolorbox}[colback=bgcolor, colframe=bordercolor, boxrule=0.5pt, arc=4pt, width=\linewidth, before=\par\vspace{6pt}, after=\par\vspace{6pt}]
\ttfamily
Please determine the appropriate caption length and then describe the actual objects in the image in a very detailed manner. Please do not include descriptions related to the focus and bokeh of this image. Please do not include descriptions like the background is blurred.
\end{tcolorbox}

\subsection{RealIR Benchmark}
In Figure~\ref{fig:realir_vis}, we present visualizations of the RealIR Benchmark. For manually captured images, we include the camera or cell phone model used at the time of shooting, along with the shooting parameters.

\begin{table*}[!t]
\resizebox{\linewidth}{!}{
\begin{tabular}{lccccccccc}
    \toprule
     \multirow{2}{*}{Methods} & \multicolumn{3}{c}{\textbf{Light Degradation}} & \multicolumn{3}{c}{\textbf{Moderate Degradation}} & \multicolumn{3}{c}{\textbf{Heavy Degradation}} \\
    \cmidrule(lr){2-4}\cmidrule(lr){5-7}\cmidrule(lr){8-10}
     & DISTS$\downarrow$ & LPIPS$\downarrow$ & LIQE$\uparrow$ & DISTS$\downarrow$ & LPIPS$\downarrow$ & LIQE$\uparrow$ & DISTS$\downarrow$ & LPIPS$\downarrow$ & LIQE$\uparrow$ \\
    \midrule
    StableSR                & 0.1791          & 0.3311          & 3.699          & 0.1864          & 0.3209          & 3.603          & 0.2181          & 0.4008          & 3.047          \\
    StableSR w/ Ours        & 0.1748          & 0.3271          & 3.733          & 0.1774          & 0.3121          & 3.872          & 0.1993          & 0.3883          & 3.502          \\
    StableSR w/ Fine-tuned Ours 
                            & \textbf{0.1641} & \textbf{0.3203} & \textbf{3.760} & \textbf{0.1652} & \textbf{0.3082} & \textbf{3.971} & \textbf{0.1918} & \textbf{0.3807} & \textbf{3.547} \\
    \bottomrule
\end{tabular}
}
\vspace{-2.8mm}
\centering
\caption{Further improvements of the fine-tuned Res-Captioner.}\label{tab:fine-tune}
\vspace{-1mm}
\end{table*}

\begin{table*}[!t]
\resizebox{0.92\linewidth}{!}{
\begin{tabular}{lccccccccc}
    \toprule
     \multirow{2}{*}{Data size (pairs)} & \multicolumn{3}{c}{\textbf{Light Degradation}} & \multicolumn{3}{c}{\textbf{Moderate Degradation}} & \multicolumn{3}{c}{\textbf{Heavy Degradation}} \\
    \cmidrule(lr){2-4}\cmidrule(lr){5-7}\cmidrule(lr){8-10}
     & DISTS$\downarrow$ & LPIPS$\downarrow$ & LIQE$\uparrow$ & DISTS$\downarrow$ & LPIPS$\downarrow$ & LIQE$\uparrow$ & DISTS$\downarrow$ & LPIPS$\downarrow$ & LIQE$\uparrow$ \\
    \midrule
    100                  & 0.1672          & 0.3212          & \textbf{3.783} & 0.1694          & 0.3092          & \textbf{3.981} & 0.1931          & 0.3815          & 3.541          \\
    200                  & 0.1641          & \textbf{0.3203} & 3.760          & \textbf{0.1652} & \textbf{0.3082} & 3.971          & 0.1918          & 0.3807          & \textbf{3.547} \\
    300                  & \textbf{0.1640} & 0.3214          & 3.742          & 0.1663          & 0.3110          & 3.967          & \textbf{0.1901} & \textbf{0.3794} & 3.529         \\
    \bottomrule
\end{tabular}
}
\vspace{-2.8mm}
\centering
\caption{Influence of fine-tuning data size.}\label{tab:fine-tune_data}
\vspace{-1mm}
\end{table*}

\section{Additional Experiments}\label{sec3}
\subsection{Res-Captioner Fine-tuning}
Although Res-Captioner performs effectively across other restoration models in a fully plug-and-play manner (see Table~\textcolor{cvprblue}{2} in the main paper), fine-tuning with a small amount of data collected on a new T2I backbone further enhances results. Table~\ref{tab:fine-tune} shows the effect of combining StableSR (based on Stable Diffusion 2.1) with a fine-tuned Res-Captioner, demonstrating improved restoration performance compared to the unfine-tuned version. This improvement is likely because fine-tuning on the new T2I backbone allows Res-Captioner to better align with the characteristics of text richness. In Table~\ref{tab:fine-tune_data}, we analyze the impact of fine-tuning data size. We find that only 200 pairs are sufficient for Res-Captioner to adapt to the new text richness rules. While increasing the data size further improves performance in heavily degraded scenarios, 200 is chosen as the optimal size considering the cost of data collection.

\begin{table*}[!t]
\resizebox{0.9\linewidth}{!}{
\begin{tabular}{lcccccccccc}
    \toprule
    \multirow{2}{*}{Methods} & \multicolumn{2}{c}{\textbf{Light Degradation}} & \multicolumn{2}{c}{\textbf{Moderate Degradation}} & \multicolumn{2}{c}{\textbf{Heavy Degradation}} & \multicolumn{2}{c}{\textbf{Average}} \\
    \cmidrule(lr){2-3}\cmidrule(lr){4-5}\cmidrule(lr){6-7}\cmidrule(lr){8-9}
     & DISTS$\downarrow$ & LPIPS$\downarrow$ & DISTS$\downarrow$ & LPIPS$\downarrow$ & DISTS$\downarrow$ & LPIPS$\downarrow$ & DISTS$\downarrow$ & LPIPS$\downarrow$ \\
    \midrule
    LLaVA-1.5            & 0.1740          & 0.3260          & 0.1764          & 0.3325          & 0.2153          & 0.4215          & 0.1886          & 0.3600          \\
    ShareCaptioner       & 0.1750          & 0.3274          & 0.1753          & 0.3171          & \textbf{0.1837} & \textbf{0.3739} & 0.1780          & 0.3394          \\
    Res-Captioner        & \textbf{0.1680} & \textbf{0.3178} & \textbf{0.1621} & \textbf{0.3052} & 0.1873          & 0.3754          & \textbf{0.1725} & \textbf{0.3328} \\
    \bottomrule
\end{tabular}
}
\vspace{-2.8mm}
\centering
\caption{Quantitative comparisons of image captioners on restoration.}\label{tab:captioner}
\vspace{-1mm}
\end{table*}

\subsection{Comparison of Image Captioners and User-defined Prompts}
We compare our Res-Captioner to LLaVA-1.5~\cite{llava-v1.5} and ShareCaptioner~\cite{share-captioner} in a plug-and-play manner, integrating all captions into SUPIR as described in Section~\textcolor{cvprblue}{2.1.1} in the main paper. Results from the multi-degradation test set, shown in Table~\ref{tab:captioner}, demonstrate that Res-Captioner offers superior guidance for image restoration when comprehensively evaluated. LLaVA-1.5 typically generates shorter captions (average length of 80), while ShareCaptioner consistently produces long captions (average length of 200). As noted in \textbf{Observation 2}, both overly short and excessively long captions can negatively affect restoration results. As a result, ShareCaptioner achieves good restoration quality only under heavy degradation, where long captions are preferred, performing slightly better than Res-Captioner in this scenario. In contrast, Res-Captioner dynamically adjusts text richness based on the input image, optimizing restoration quality across varying degradation levels. 

Figures~\ref{fig:captioner_vis} and \ref{fig:captioner_vis_1} present a comparison of captions generated by different image captioners for the same low-quality (LQ) images and their corresponding restoration effects. Additionally, we include manually designed captions (User-defined) for comparison. The results show that captions generated by LLaVA-1.5 and ShareCaptioner often include hallucinations and harmful descriptions, and their lengths are relatively fixed. While user-defined captions are generally accurate, they tend to be shorter when provided by non-professional annotators. In contrast, Res-Captioner generates captions that avoid harmful descriptions, include fewer hallucinations, and adaptively adjust length based on the LQ image. This adaptability leads to the best restoration quality among the evaluated methods.

\begin{table*}[!t]
\resizebox{\linewidth}{!}{
\begin{tabular}{lcccccccccccc}
    \toprule
     \multirow{2}{*}{Methods} & \multicolumn{4}{c}{\textbf{Light Degradation}} & \multicolumn{4}{c}{\textbf{Moderate Degradation}} & \multicolumn{4}{c}{\textbf{Heavy Degradation}} \\
    \cmidrule(lr){2-5}\cmidrule(lr){6-9}\cmidrule(lr){10-13}
      & DISTS$\downarrow$ & LPIPS$\downarrow$ & MANIQA$\uparrow$ & LIQE$\uparrow$ & DISTS$\downarrow$ & LPIPS$\downarrow$ & MANIQA$\uparrow$ & LIQE$\uparrow$ & DISTS$\downarrow$ & LPIPS$\downarrow$ & MANIQA$\uparrow$ & LIQE$\uparrow$ \\
    \midrule
    CoSeR                & 0.1880          & 0.3322          & 0.1159          & 2.670          & 0.1978          & 0.3405          & 0.1203          & 2.699          & 0.2208          & 0.4019          & 0.0952          & 2.235          \\
    CoSeR w/ Ours        & \textbf{0.1787} & \textbf{0.3216} & \textbf{0.1360} & \textbf{3.101} & \textbf{0.1876} & \textbf{0.3327} & \textbf{0.1350} & \textbf{3.194} & \textbf{0.2187} & \textbf{0.4005} & \textbf{0.0981} & \textbf{2.409} \\
    \bottomrule
\end{tabular}
}
\vspace{-2.8mm}
\centering
\caption{Quantitative comparisons of our method applied to CoSeR.}\label{tab:coser_comparison}
\vspace{-1mm}
\end{table*}





\begin{figure*}[htbp]
    \centering
    \begin{subfigure}[t]{\textwidth}
        \centering
        \includegraphics[width=\textwidth]{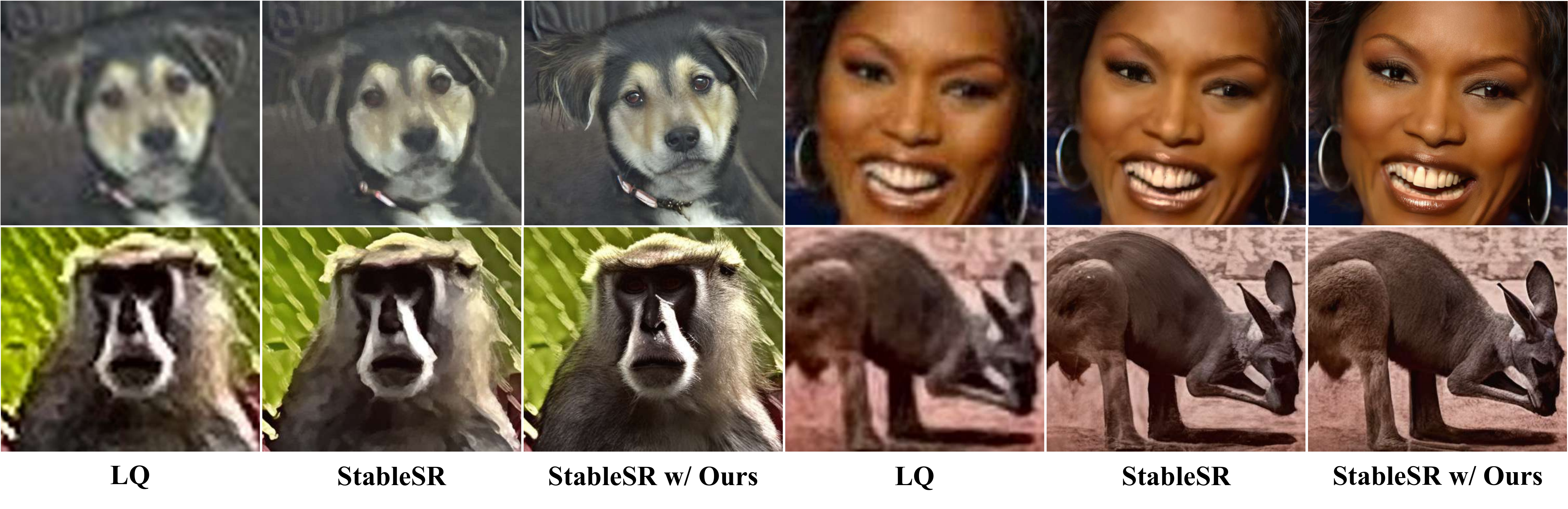}
        \vspace{-5mm}
        \caption{Additional qualitative comparisons of Res-Captioner applied to StableSR on in-the-wild images.}
        \label{fig:supp_comp_2}
    \end{subfigure}
    
    \begin{subfigure}[t]{\textwidth}
        \centering
        \includegraphics[width=\textwidth]{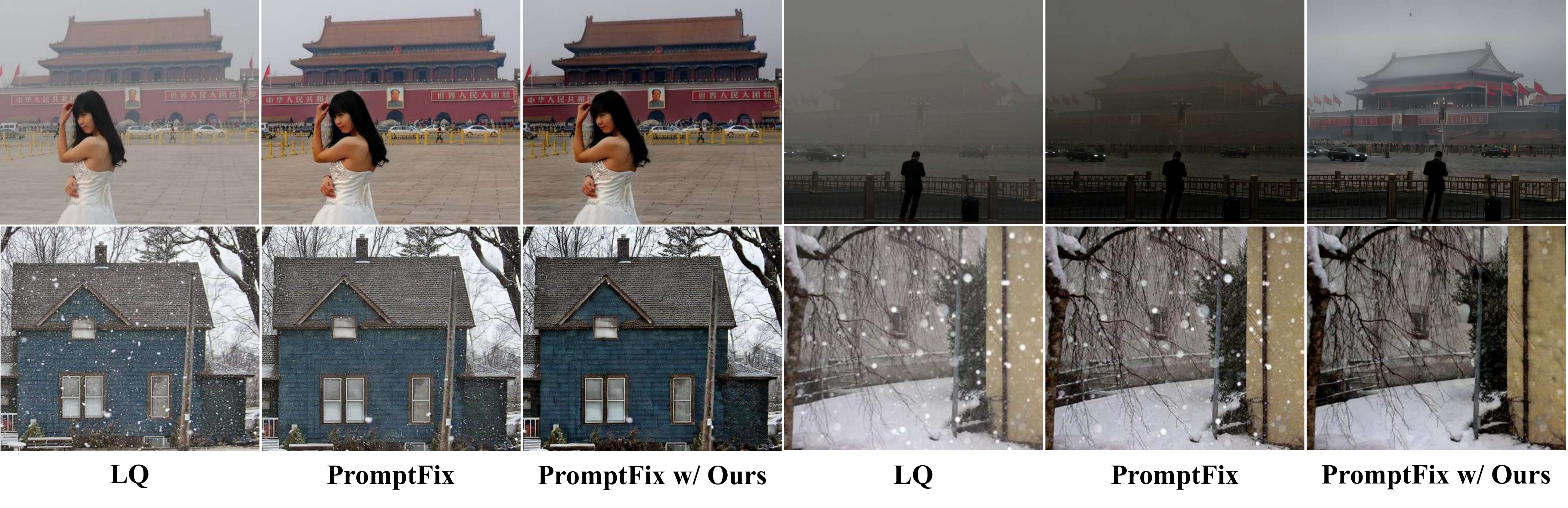}
        \vspace{-5mm}
        \caption{Qualitative comparisons of Res-Captioner on de-hazing and de-snowing.}
        \label{fig:dehaze_desnow}
    \end{subfigure}
    
    \caption{Qualitative comparisons of Res-Captioner on various tasks.}
    \label{fig:combined_figure}
\end{figure*}

\subsection{Quantitative Comparisons on CoSeR}
Table~\ref{tab:coser_comparison} illustrates the impact of combining Res-Captioner with CoSeR~\cite{coser}, further demonstrating the generalization ability of Res-Captioner.

\subsection{More Qualitative Results}
In this section, we provide additional visual comparisons between our method and state-of-the-art (SOTA) methods. As illustrated in Figure~\ref{fig:supp_comp}, when paired with the SUPIR restoration method, which features a powerful generative model backbone, our Res-Captioner shows clear advantages in recovery performance compared to previous SOTA methods. Additionally, our approach significantly improves visual quality when applied to the StableSR restoration method, as proved in Figure~\ref{fig:supp_comp_2}, highlighting the robustness of our method across different restoration models. 

Figure~\ref{fig:dehaze_desnow} illustrates the effectiveness of our method on specific image restoration tasks, including de-hazing and de-snowing, when combined with the all-in-one image restoration model PromptFix~\cite{promptfix}. The inputs are sourced from the real-world de-hazing dataset~\cite{rtts} and the real-world de-snowing dataset~\cite{snow100k}.

\subsection{Analysis on Degradation Token Number}
We analyze the impact of the token length $N$ in our degradation-aware visual encoder on text richness prediction accuracy. As shown in Table~\ref{tab:deg_token}, the token length is inversely proportional to the text richness offset level $E$, indicating a positive relationship with text richness prediction accuracy. Balancing computational cost with performance benefits, we select $N=36$ as the final token number.

\begin{table}[h]
\centering

\begin{tabular}{c|c}
\toprule
Token Length $N$ & Offset Level $E$ \\ \midrule
4                              & 1.53             \\
9                              & 1.46             \\
16                             & 1.34             \\
25                             & 1.28             \\
\textbf{36}                    & \textbf{1.27}    \\ \bottomrule
\end{tabular}
\caption{Analysis of the token length in the degradation-aware visual encoder.}\label{tab:deg_token}
\end{table}

\begin{figure*}[htbp]
    \centering
    \includegraphics[width=\textwidth]{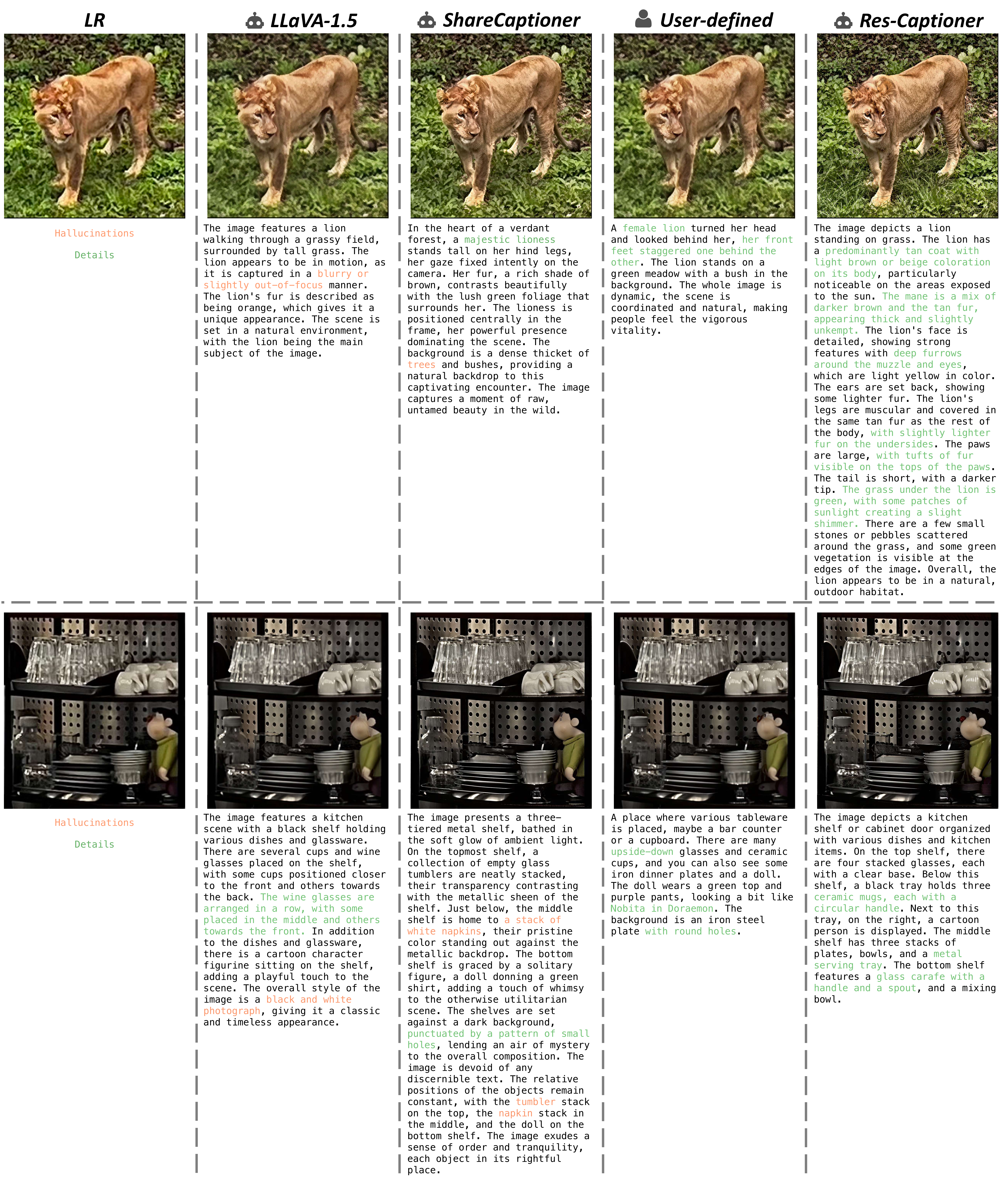}
    \vspace{-7mm}
    \caption{Comparison of restoration performance for different image captioners and user-defined prompts.}
    \label{fig:captioner_vis}
\end{figure*}

\begin{figure*}[htbp]
    \centering
    \includegraphics[width=\textwidth]{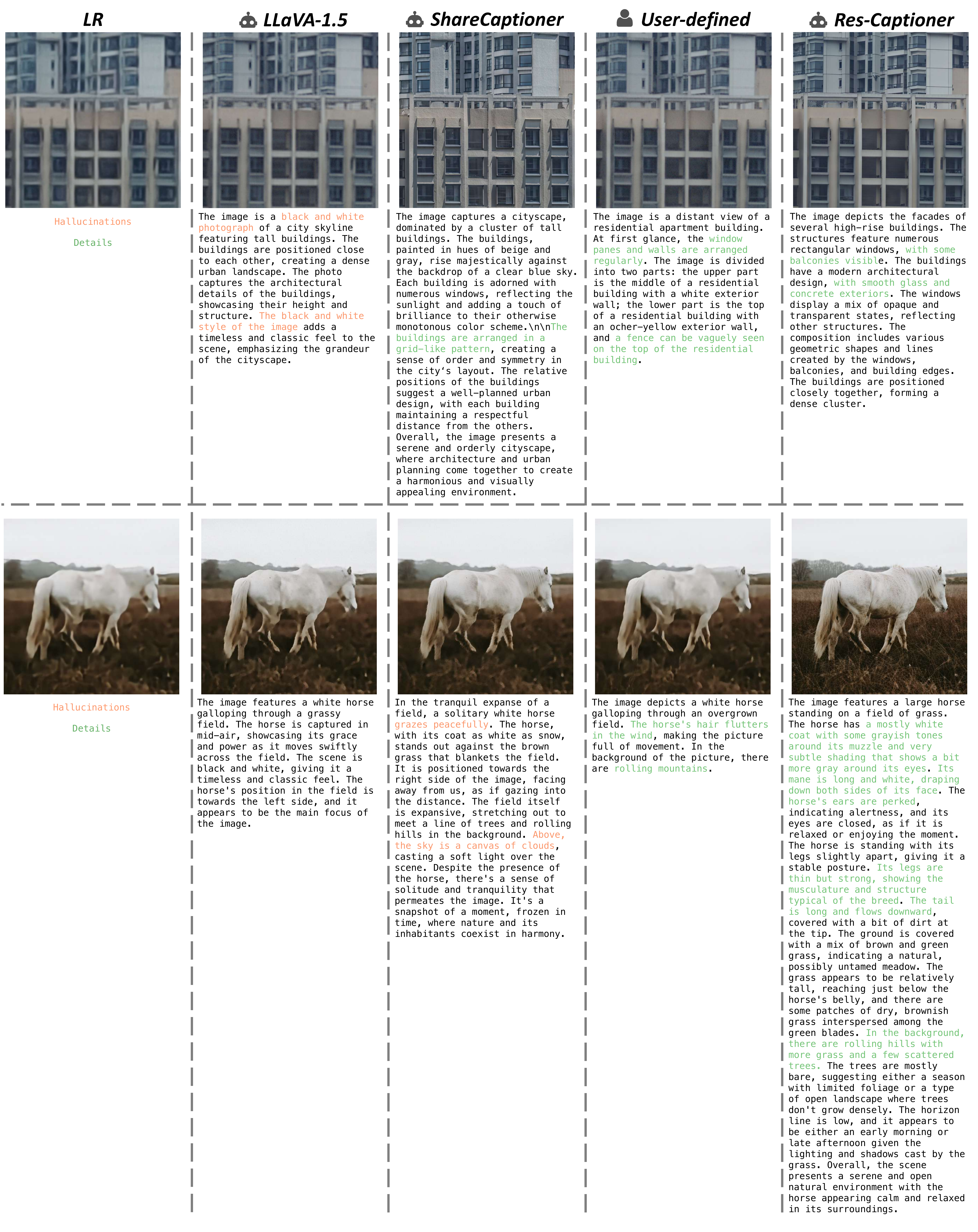}
    \vspace{-7mm}
    \caption{Comparison of restoration performance for different image captioners and user-defined prompts.}
    \label{fig:captioner_vis_1}
\end{figure*}

\begin{figure*}[htbp] 
    \centering
    \includegraphics[width=\textwidth]{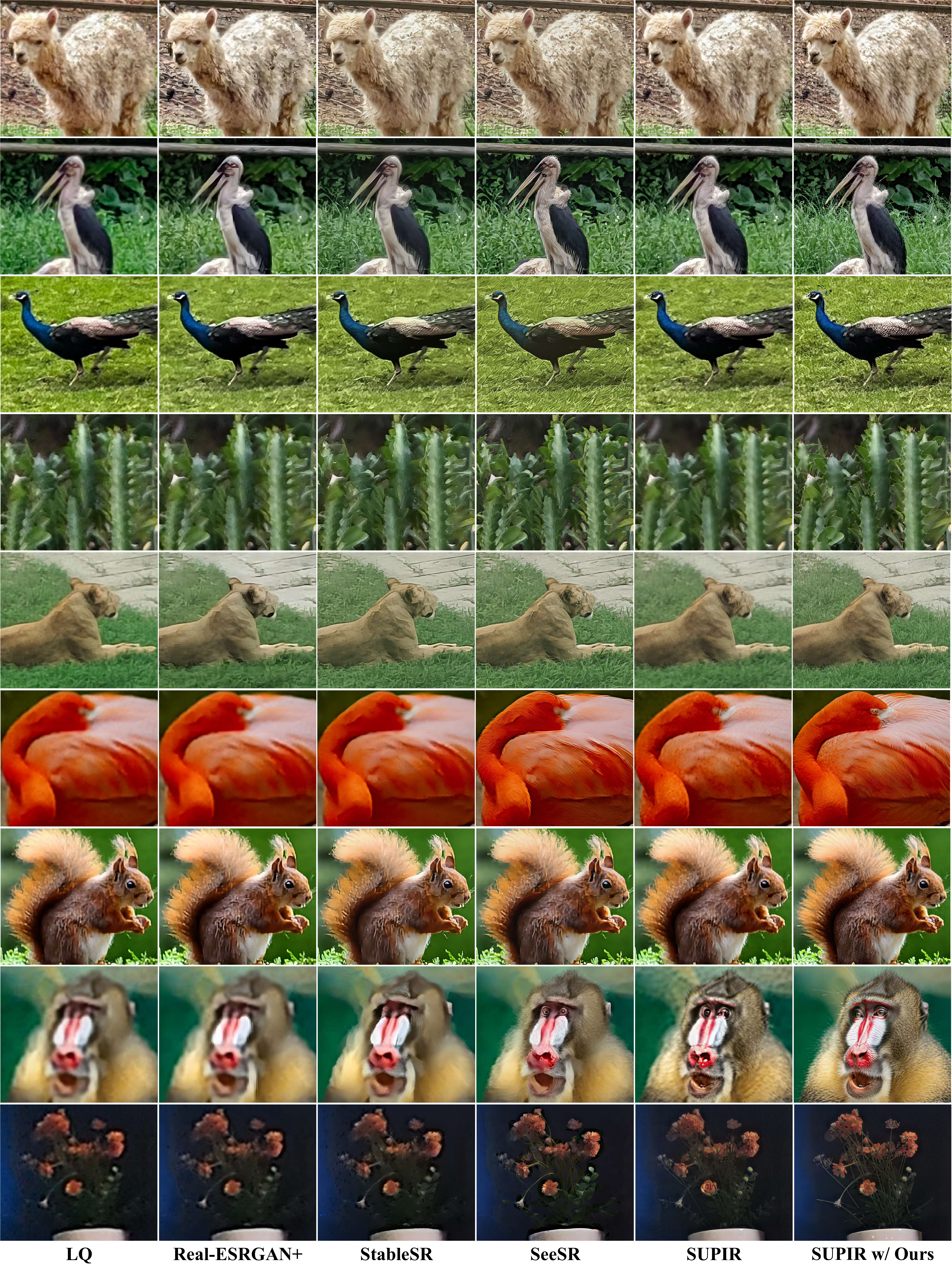}
    \caption{Additional qualitative comparisons with SOTA methods on in-the-wild images.}
    \label{fig:supp_comp}
\end{figure*}

\section{Limitations and Future Work}\label{sec4}
SeeSR~\cite{seesr} and CoSeR~\cite{coser} utilize intermediate representations containing semantic information as conditional inputs. This information includes not only abstract semantic details but also fine-grained image features, which may be beneficial for robust restoration. This type of information has not yet been fully exploited in our current model. In the future, we plan to explore its role as invariant representations for restoration models.

\clearpage
\clearpage

{
    \small
    \bibliographystyle{ieeenat_fullname}
    \bibliography{main}
}


\end{document}